\documentclass[a4paper, 10 pt, conference]{ieeeconf}
\IEEEoverridecommandlockouts                              
\usepackage{graphicx} 
\usepackage{epsfig} 
\usepackage{mathptmx} 
\usepackage{times} 
\usepackage{amsmath} 
\usepackage{amssymb} 
\usepackage{multirow}
\usepackage{url}
\usepackage{bm}
\usepackage{booktabs}
\usepackage{caption}

\begin{document}

\title{\LARGE \bf
Bi-MoDe: Bilateral Control-based Imitation Learning via Modifier-Conditioned Decoding for Modulation of \\ Execution Speed and Contact Intensity
}

\author{Takumi Kobayashi$^{1\dag}$, Masato Kobayashi$^{1,2,3\dag*}$, Yuki Uranishi$^{1,2}$ 
\thanks{
${\dag}$ Equal Contribution,
$^{1}$ Graduate School of Information Science and Technology, The University of Osaka, $^{2}$ D3 Center, The University of Osaka, $^{3}$ Graduate School of Maritime Sciences, Kobe University, * corresponding author: kobayashi.masato.cmc@osaka-u.ac.jp}
}

\maketitle
\begin{abstract}
Bilateral control-based imitation learning captures both position and force information, making it well suited to contact-rich manipulation.
However, existing approaches provide limited means for an operator to specify how a learned task should be executed at inference time, such as slowly or quickly, gently or firmly.
We propose Bi-MoDe, a modifier-conditioned decoding framework that injects a constrained latent into every layer of the Transformer action decoder via adaLN-Zero, allowing behavioral directives to directly influence action-chunk generation.
We evaluate the method on a real-world whiteboard wiping task with combinations of temporal and physical modifiers.
Bi-MoDe improves physical directive following over the action-chunking baseline while maintaining comparable temporal control.
An ablation further shows that decoder conditioning and latent-space composition interact, and that their combination is important for accurate physical directive following.
Additional material is available at the \url{https://mertcookimg.github.io/bi-mode/}
\end{abstract}

\section{INTRODUCTION}

Imitation learning has emerged as a promising approach for enabling robots to acquire complex manipulation skills directly from human demonstrations, without the need for explicit programming or reward engineering~\cite{zhao2023learning, chi2025diffusion}.
By observing and replicating human behavior, robots can learn to perform a wide range of tasks in unstructured environments, making imitation learning applicable across industrial, service, and general-purpose manipulation domains.

\begin{figure}[t]
\centering
\includegraphics[keepaspectratio, width=\linewidth]{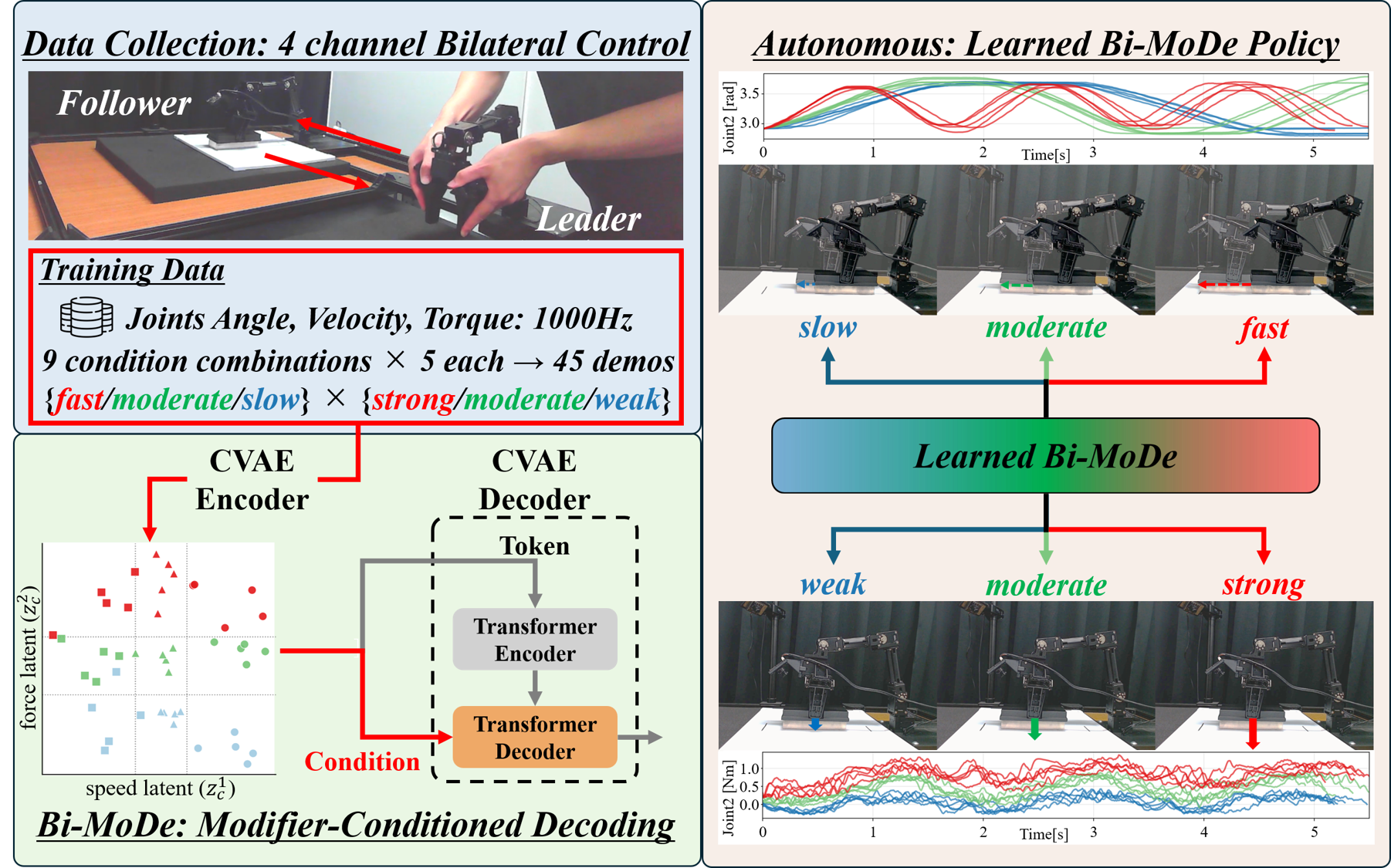}
\caption{Concept of Bi-MoDe }
\label{fig:teaser}
\end{figure}

A task, however, is rarely specified completely by its name.
The same wiping motion must be executed gently on a fragile surface and firmly on a stubborn stain, quickly when time is scarce and slowly when precision matters.
These are properties of the execution rather than of the task, and a policy that has learned only the average of its demonstrations cannot be asked to shift along them.
Encoding such behavioral variability into a learned policy, and exposing it as something an operator can specify at inference time, remains a non-trivial problem.
Fig.~\ref{fig:teaser} shows what this amounts to on the task studied here: the same wiping motion executed at a commanded speed and a commanded contact force.

Bilateral control-based imitation learning is well positioned to address the physical half of this problem.
By simultaneously controlling position and force on both the leader and follower robots, this approach records the contact forces that unilateral teleoperation discards, and it reproduces contact-rich manipulation skills faithfully~\cite{adachi2018imitation, buamanee2024biact}.
Execution speed has been studied within this framework, through policies trained on variable-speed contact motion~\cite{sakaino2022imitation} and through reference trajectories refined so that demonstrations can be replayed faster~\cite{yamane2026refinement}.
In these approaches the desired behavior is engineered into the training data before the policy is learned, so the behavioral characteristics of the resulting trajectory remain a property of the dataset rather than a quantity the operator can vary once the policy is deployed.

\begin{figure*}[t]
\centering
\includegraphics[keepaspectratio, width=1.0\linewidth]{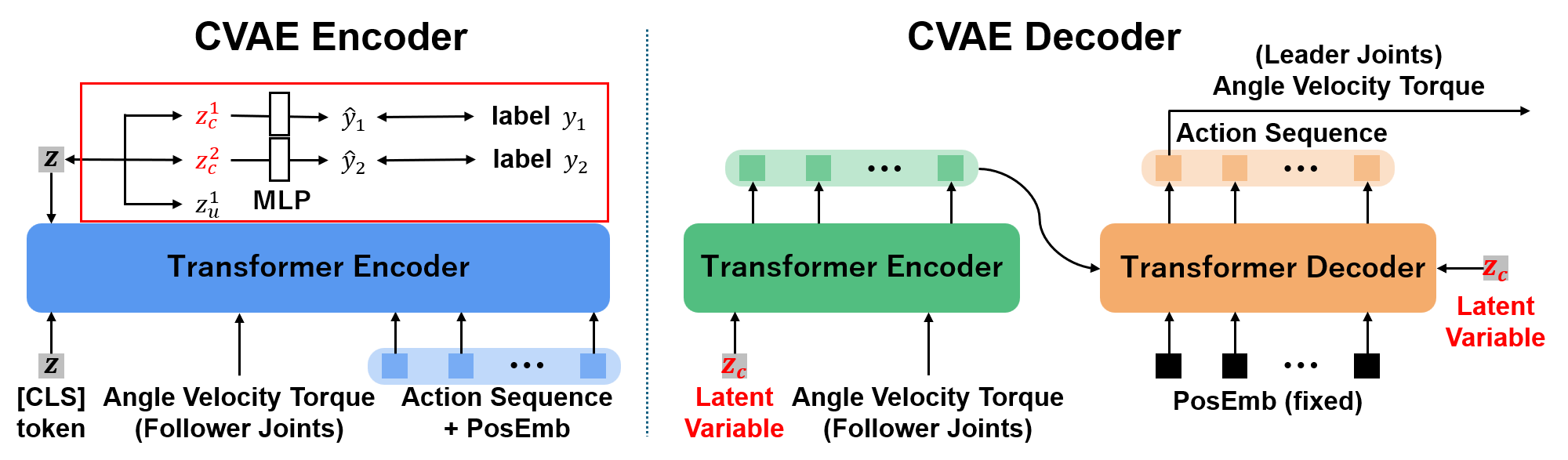}
\caption{Overview of Bi-MoDe. The constrained latent $\bm{z}_c$ is appended as a token to the Transformer encoder input and, through adaLN-Zero, modulates every layer of the action decoder.}
\label{fig:model}
\end{figure*}

Several interfaces for such specification have been proposed, each with a different failure.
Natural language is intuitive but discrete and ambiguous, and an abstract phrase does not map to a reproducible force magnitude~\cite{kobayashi2025bilat, robledo2026tacstyle}.
Scalar conditioning is quantitative, but it has so far been applied to the temporal axis alone~\cite{jing2026tempovla}.
Closest to our setting, Oishi et al.~\cite{oishi2025imitation} introduced modifier directives, structured scalar labels encoding the desired temporal and physical characteristics of each demonstration, and proposed a disentangled latent representation that aligns constrained latent dimensions to modifier targets via weakly supervised labels.
This approach makes behavioral control quantitative and reproducible.
Their formulation was evaluated with two sequence backbones, an LSTM and an action chunking Transformer, and two limitations remained.
Interference between the two behavioral characteristics persisted, and the action chunking variant followed the commanded directives less reliably than the LSTM.
We address the second of these, and take the action chunking architecture as our scope, since it is the backbone on which recent bilateral control-based policies are built and the one whose directive following is least understood.

We argue that this limitation is not a property of modifier directives but of where the condition is allowed to act.
In~\cite{oishi2025imitation} the modifier is conditioned only at the encoder stage, so the action decoder generates trajectories without explicit access to the directive and must recover it from a weakly constrained latent variable.
Regularizing that latent toward a prior does not make it clean, since matching a marginal distribution places no constraint on the distribution conditioned on the directive~\cite{bach2026marginal}.
This is particularly consequential for an action chunking decoder, whose behavioral intent must persist across an entire chunk rather than modulate a single prediction.
Where a condition should enter a Transformer policy has been studied for observations and diffusion timesteps~\cite{dasari2025ingredients, yan2025maniflow} but not for this kind.
In this paper, we formulate behavioral specification as a conditioning design problem.
We introduce Bi-MoDe, a bilateral control-based imitation learning framework with
modifier-conditioned decoding via adaLN-Zero~\cite{peebles2023scalable}.
Bi-MoDe injects the constrained latent $\bm{z}_c$ into every Transformer decoder layer,
allowing temporal and physical modifier directives to directly influence action-chunk
generation rather than being conveyed only through the encoded observation.
This design enables the learned policy to modulate execution speed and contact intensity
according to operator-specified directives at inference time.

The contributions of this paper are as follows:
\begin{itemize}
    \item We propose Bi-MoDe, which injects the constrained latent
    $\bm{z}_c$ into every Transformer action-decoder layer via adaLN-Zero
    to directly condition action-chunk generation.

    \item We demonstrate inference-time modulation of execution speed and
    contact intensity on a real-world contact-rich wiping task.

    \item A $2\times2$ ablation shows that decoder conditioning and
    latent-space composition jointly improve physical directive following.
\end{itemize}

\section{RELATED WORK}

\subsection{Bilateral Control-based Imitation Learning}

Bilateral control enables leader-follower teleoperation that captures both position and force information, making it well suited to contact-rich manipulation~\cite{adachi2018imitation, hayashi2022independently, tsuji2026survey}.
Bi-ACT extended Action Chunking with Transformers to this framework~\cite{buamanee2024biact, kobayashi2025alpha}, followed by studies on data augmentation~\cite{kobayashi2025dabi}, vision-language fusion~\cite{kobayashi2026bivla}, environmental adaptation~\cite{tsunoori2025biaqua}, and behavioral specification through modifier directives~\cite{oishi2025imitation}.

Execution speed has also been addressed through variable-speed contact motion~\cite{sakaino2022imitation}, teaching-playback augmentation~\cite{masuya2025variable}, and accelerated trajectory refinement~\cite{yamane2026refinement}.
However, these approaches encode behavioral characteristics into demonstrations or reference trajectories before training rather than exposing them as inference-time commands.
Contact-rich wiping has likewise been studied under varying surface conditions~\cite{tsuji2025adaptive}, but these variations describe the environment rather than operator-specified behavior.
Bi-MoDe instead enables temporal and physical characteristics to be specified at inference time through action-decoder conditioning.

\subsection{Behavior-Level Conditioning of Manipulation Policies}

Behavior-level conditioning controls not only \emph{what} a robot does but also \emph{how} it performs a task.
Natural-language conditioning has been used to modulate force in bilateral imitation learning~\cite{kobayashi2025bilat}, but linguistic instructions remain ambiguous for specifying continuous physical quantities~\cite{robledo2026tacstyle}.
Scalar conditioning provides a quantitative alternative: TempoVLA controls execution speed~\cite{jing2026tempovla}, while adaptive compliance policies regulate physical interaction autonomously~\cite{hou2025adaptive}; neither directly provides operator-specified control over both temporal and physical characteristics.

Closest to our work, modifier directives align constrained latent dimensions with temporal and physical labels through weak supervision~\cite{oishi2025imitation}.
This formulation builds on latent subspace learning~\cite{klys2018learning} and weakly supervised disentanglement~\cite{locatello2020weakly}, but its action chunking variant follows commanded directives less reliably than its LSTM counterpart.
Because conditioning is confined to the encoder, the decoder must recover the directive from a weakly constrained latent representation, whose marginal regularization does not guarantee condition-specific separation~\cite{bach2026marginal}.
Bi-MoDe therefore retains modifier directives while propagating the constrained latent directly through the action decoder.

\section{METHOD}
\subsection{Overview}\label{sec:overview}
Bi-MoDe extends~\cite{oishi2025imitation} by introducing \textit{modifier-conditioned decoding} via Adaptive Layer Normalization with zero initialization (adaLN-Zero)~\cite{peebles2023scalable}, as illustrated in Fig.~\ref{fig:model}.
We take the action chunking Transformer as the backbone throughout, since it is the variant for which~\cite{oishi2025imitation} report the least reliable directive following, and the architecture on which recent bilateral control-based policies build~\cite{buamanee2024biact, kobayashi2025bilat, tsunoori2025biaqua}.
By propagating the constrained latent $\bm{z}_c$ into every Transformer layer of the decoder, in addition to the encoder token that both methods share,
Bi-MoDe allows modifier directives to directly influence
the trajectory generation process, enabling more precise and consistent behavioral modulation.

\subsection{Data Collection}\label{sec:datacollection}

Bi-MoDe employs the four-channel bilateral control method for teleoperation data collection, following~\cite{oishi2025imitation}. Bilateral control achieves coordinated leader-follower behavior by continuously synchronizing positional and force information between the operator and the robot.
The governing constraints are:
\begin{equation}
    \theta_l - \theta_f = 0
\end{equation}
\begin{equation}
    \tau_l + \tau_f = 0
\end{equation}
where $\theta$ and $\tau$ denote joint angles and torques, and the subscripts $l$ and $f$ refer to the leader and follower systems, respectively.
Joint angles are measured via encoders, while torque responses are estimated using a disturbance observer (DOB) and a reaction force observer (RFOB), eliminating the need for dedicated force/torque sensors.

The data collection protocol follows~\cite{oishi2025imitation}: prior to each demonstration, the operator assigns a pair of modifier labels $(y_\text{temp},\, y_\text{phys}) \in \{0.0,\, 0.5,\, 1.0\}^2$ specifying the intended temporal and physical characteristics of that trial.
A \textit{temporal modifier} governs motion speed ($\text{slow}=0.0$, $\text{moderate}=0.5$, $\text{fast}=1.0$) and a \textit{physical modifier} governs contact force ($\text{weak}=0.0$, $\text{moderate}=0.5$, $\text{strong}=1.0$).
These scalar labels are recorded synchronously with joint state trajectories.
The label is assigned before the trial rather than inferred afterwards, so the operator adjusts the motion to the intended level rather than describing what was done.

\subsection{Learning Model}\label{sec:model}
Bi-MoDe is built upon the Transformer-driven Conditional Variational Autoencoder (CVAE) architecture of~\cite{oishi2025imitation}, which partitions the latent vector into a constrained component $\bm{z}_c$ and an unconstrained component $\bm{z}_u$, as shown in Fig.~\ref{fig:model}. The CVAE encoder on the left estimates the latent variable from the demonstrated action sequence and the follower state, and a weakly supervised head aligns $\bm{z}_c$ with the modifier labels.

The remainder of the model receives the joint angle, velocity, and torque data of the follower robot as input and outputs action chunks specifying the predicted joint angles, velocities, and torques of the leader robot.
In Bi-MoDe, the unconstrained latent $\bm{z}_u$ is still inferred by the CVAE encoder and regularized by the KL term during training, but it is excluded from the latent token, which therefore contains only the constrained latent $\bm{z}_c$.

The constrained latent $\bm{z}_c$ enters the model through two conditioning paths.
As in~\cite{oishi2025imitation}, it is projected and appended as a token to the Transformer encoder input together with the proprioceptive state.
Bi-MoDe additionally projects $\bm{z}_c$ to modulate every layer of the action decoder via adaLN-Zero, as described below.
This design retains the original encoder-side conditioning while providing the decoder with direct access to the behavioral directive throughout action-chunk generation.

\begin{figure}[t]
\centering
\includegraphics[keepaspectratio, width=1.0\linewidth]{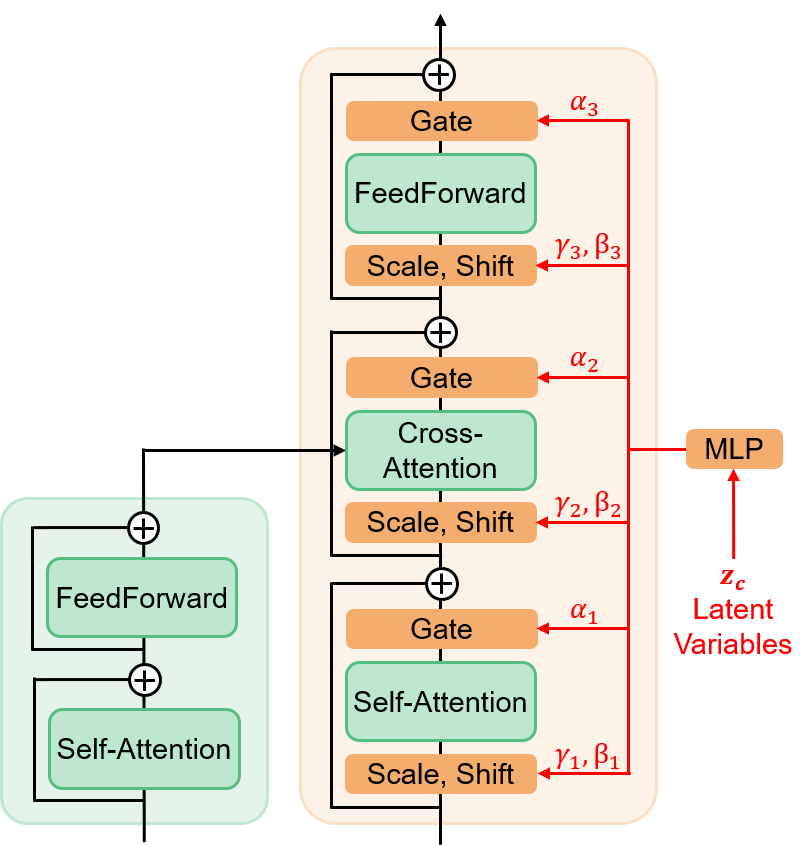}
\caption{One encoder layer and one decoder layer. A zero-initialized MLP regresses the scale, shift and gate of all three decoder sub-layers from $\bm{z}_c$.}
\label{fig:adaln}
\end{figure}

The key contribution of Bi-MoDe is the introduction of \textit{modifier-conditioned decoding} via adaLN-Zero~\cite{peebles2023scalable}.
Rather than relying solely on the encoder-stage disentanglement of~\cite{oishi2025imitation}, the constrained latent $\bm{z}_c$ is injected into every Transformer decoder layer, where it modulates each of the three sub-layers, namely self-attention, cross-attention, and the feedforward network. Fig.~\ref{fig:adaln} contrasts one such layer with an encoder layer, which carries no modulation. Writing $F_i$ for the $i$-th sub-layer and $\mathrm{LN}$ for a LayerNorm without affine parameters, the update is given by Eq.~\eqref{eq:adaln}:
\begin{equation}
    \bm{h} \leftarrow \bm{h} + (1 + \alpha_i)\,
    F_i\bigl( (1 + \gamma_i)\,\mathrm{LN}(\bm{h}) + \beta_i \bigr)
    \label{eq:adaln}
\end{equation}
For the cross-attention the encoder output supplies the keys and values, so the modulation applies to the queries alone.
The scale $\gamma_i$, shift $\beta_i$, and gate $\alpha_i$ of all three sub-layers are regressed from $\bm{z}_c$ by the single MLP drawn at the right of Fig.~\ref{fig:adaln}, whose output layer is zero-initialized, so that all three vanish for any $\bm{z}_c$ at the start of training.
At initialization, each layer therefore applies no direct adaLN modulation, while conditioning through the encoder-side latent token remains active via the cross-attention memory.

Gating the residual branch by $(1 + \alpha_i)$ rather than by $\alpha_i$
departs from the usual formulation, in which the gate is applied directly so that each block is initialized as an identity function~\cite{peebles2023scalable}.
With a nonzero block input, such as an image patch or a noised action token, identity initialization preserves that input in the residual stream.
The DETR-style decoder used here initializes its residual stream to zero and supplies the position of each action step separately as a learned embedding, so observation-dependent information enters through the sub-layer branches.
Gating those branches by $\alpha_i$ alone would close all of them at initialization, leaving the decoder hidden output zero and the predicted action equal to the output head's bias.
The gates could subsequently learn to open, but using $(1 + \alpha_i)$ allows observation-dependent information to pass through the residual branches from initialization.
Zero initialization thus gives unit residual gates and zero direct adaLN modulation, while retaining conditioning through the encoder memory.

The motivation for zero-initializing the direct modulation is that the conditioning variable $\bm{z}_c$ is itself learned under weak supervision.
Its alignment with the modifier labels is not established at initialization, so zero initialization avoids applying a randomly initialized adaLN transformation of this evolving representation.
The latent representation and decoder modulation are then learned jointly under the training objective below.
The modulation output layer can update from the first optimization step, and the encoder-side conditioning remains active throughout; no ordering in which latent alignment must precede decoder conditioning is imposed.
This design provides a learned direct conditioning path to every decoder layer without an explicit schedule for enabling modulation.

The training objective follows that of~\cite{oishi2025imitation}.
We replace the squared-error reconstruction loss with an $\ell_1$ loss for action-chunk prediction (Eq.~\eqref{eq:rec}), and retain the KL regularization (Eq.~\eqref{eq:kl}) and modifier prediction losses from the original formulation.
For each of the $S$ constrained latent dimensions, $S=2$ here, an MLP head maps the corresponding scalar $z_{c,s}$ to a logit $\hat{y}_s = \mathrm{MLP}(z_{c,s})$, following the architecture of~\cite{oishi2025imitation}.
The modifier prediction loss $L_{\mathrm{modi}}$ accumulates a binary cross-entropy against the label $y_s \in \{0.0, 0.5, 1.0\}$ over all such dimensions (Eq.~\eqref{eq:modi}):
\begin{equation}
    L_{\mathrm{rec}} = \frac{1}{k} \sum_{j=0}^{k-1}
    \bigl| \bm{a}_{t+j} - \hat{\bm{a}}_{t+j} \bigr|
    \label{eq:rec}
\end{equation}
\begin{equation}
    L_{\mathrm{kl}} =
    D_{\mathrm{KL}}\bigl[
        q_p(\bm{z} \mid \bm{a}_{t:t+k},\, \bm{s}_t)
        \,\|\,
        \mathcal{N}(0, \bm{I})
    \bigr]
    \label{eq:kl}
\end{equation}
\begin{equation}
    L_{\mathrm{modi}} = -\sum_{s=1}^{S} \Bigl[
        y_s \cdot \log\bigl(\sigma(\hat{y}_s)\bigr)
        + (1 - y_s) \cdot \log\bigl(1 - \sigma(\hat{y}_s)\bigr)
    \Bigr]
    \label{eq:modi}
\end{equation}

The encoder, the modifier prediction head, and the decoder are optimized under Eq.~\eqref{eq:total}, with the loss weights given in Table~\ref{tab:hyperparams}.
\begin{equation}
    L = \lambda_{\mathrm{rec}} L_{\mathrm{rec}}
      + \lambda_{\mathrm{kl}} L_{\mathrm{kl}}
      + \lambda_{\mathrm{modi}} L_{\mathrm{modi}}
    \label{eq:total}
\end{equation}

\begin{figure}[t]
\centering
\includegraphics[keepaspectratio, width=1.0\linewidth]{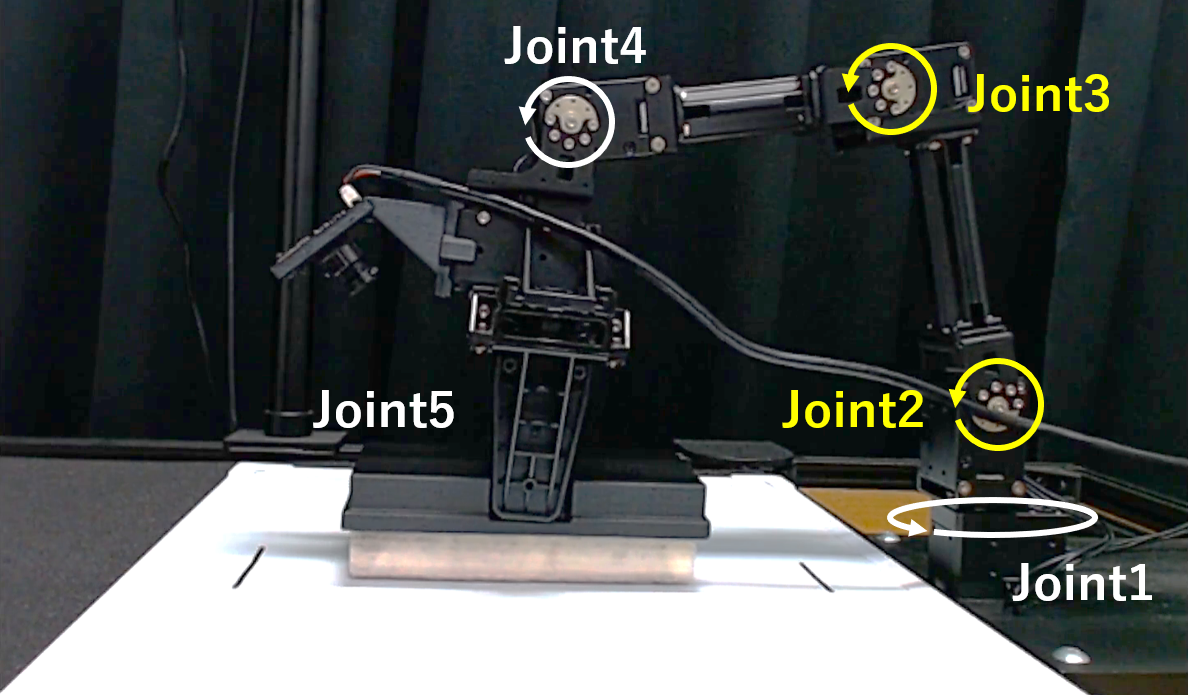}
\caption{The follower robot, with Joint2 and Joint3 highlighted.}
\label{fig:robot}
\end{figure}

\subsection{Inference}\label{sec:inference}

During inference, Bi-MoDe receives the most recent joint data from the follower robot.
The operator specifies behavior by selecting a directive level on each axis, and each level is realized by a fixed value of $\bm{z}_c$ determined in advance from the training data: the trained encoder is applied to every demonstration, and the median of the resulting constrained latents within each label level is taken as the command for that level.
These values are properties of the learned latent space rather than of the directive scale, so they differ between models and their sign is arbitrary; what the procedure guarantees is that a given directive level is realized by the same latent command on every rollout.
Based on these inputs, the model predicts the subsequent action chunk of the leader robot's joint information, with modifier directives propagated throughout the decoder via adaLN-Zero conditioning at every layer.

\begin{figure*}[t]
\centering
\includegraphics[keepaspectratio, width=1\linewidth]{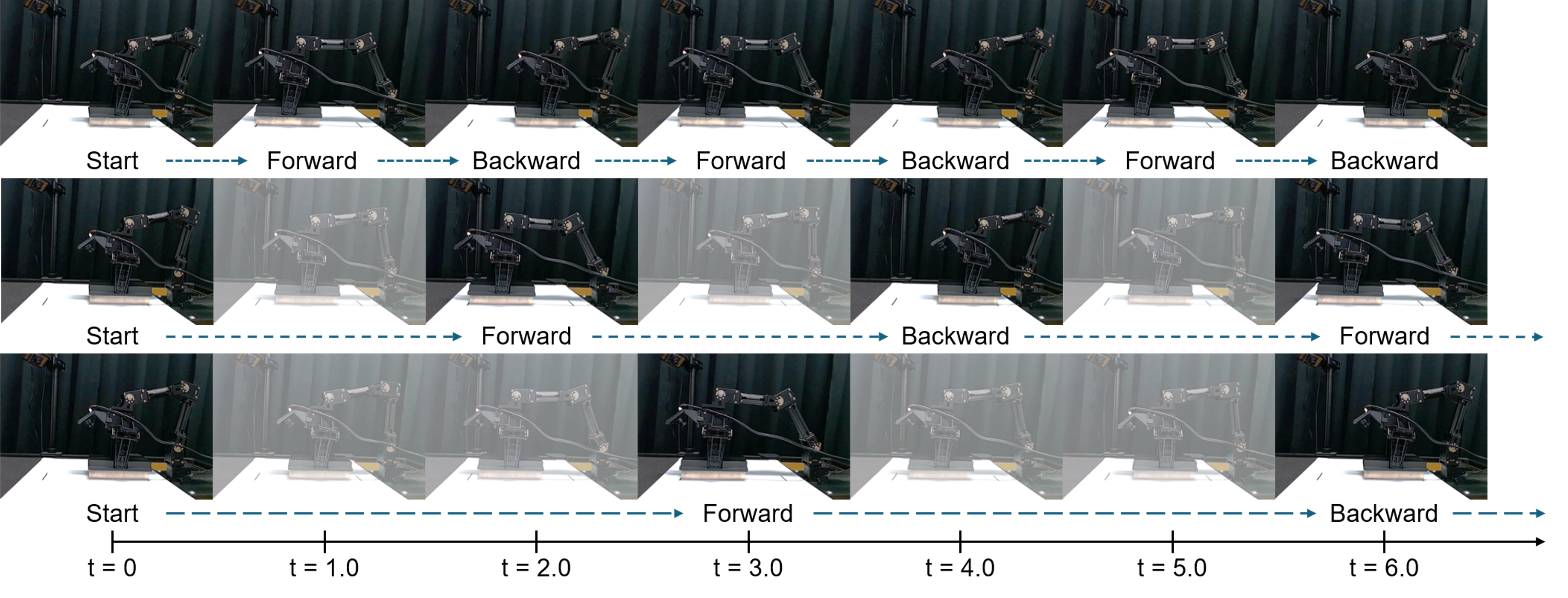}
\caption{Effect of the temporal modifier in the demonstrations. Rows are fast, moderate and slow from top to bottom, with frames taken at equal elapsed time: over six seconds the three conditions complete three, one and a half, and one stroke.}
\label{fig:task}
\end{figure*}

\begin{figure}[t]
\centering
\includegraphics[keepaspectratio, width=1\linewidth]{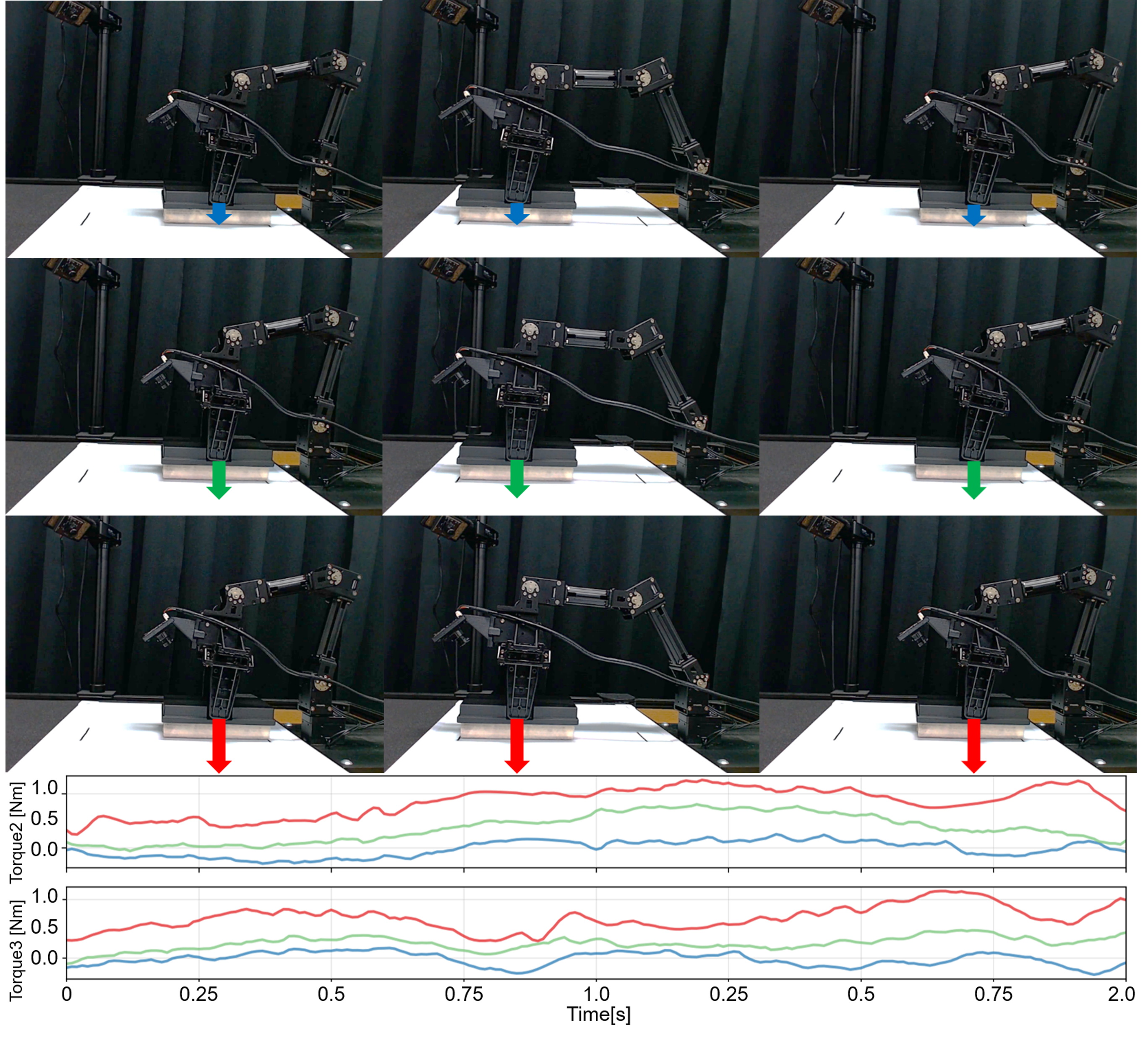}
\caption{Effect of the physical modifier in the demonstrations. Rows are weak, moderate and strong, with columns showing the start, forward and backward phases of one stroke. The arm follows the same path at every level, while the Joint2 and Joint3 torques below separate by commanded force.}
\label{fig:force}
\end{figure}

\section{EXPERIMENT}

\subsection{Hardware}
We used the OpenManipulator-X from ROBOTIS, which features four joints (Joint1 to Joint4) for arm rotation and Joint5 for gripper actuation, with two arms serving as the leader and follower robots.
Joint2 and Joint3, highlighted in Fig.~\ref{fig:robot}, bear the load when the cleaner is pressed against the board, and their torque is used as the physical feature in the evaluation.

\subsection{Task Setting}
We evaluated Bi-MoDe on a whiteboard wiping task using a whiteboard cleaner.
The follower robot holds the cleaner against the board and sweeps from one end to the other and back, and this stroke is repeated three times.
The task is well suited to evaluating modifier directive control, as both modifiers manifest as distinct and measurable characteristics in the demonstrations.
Fig.~\ref{fig:task} shows the effect of the temporal modifier: frames taken at equal elapsed time reveal that a faster command completes more strokes within the same interval.
Fig.~\ref{fig:force} shows the effect of the physical modifier: the arm posture at each phase of a stroke is nearly identical across commanded levels, while the torques of Joint2 and Joint3, which bear the load during contact, separate by level throughout.

Demonstrations were collected for all combinations of temporal and physical modifiers, $\{\text{fast},\, \text{moderate},\, \text{slow}\} \times \{\text{strong},\, \text{moderate},\, \text{weak}\}$, yielding 9 conditions.
5 demonstrations were recorded per condition, for a total of 45 demonstrations.
At inference, each configuration was evaluated over the same 9 conditions with 5 rollouts per condition, for 45 rollouts per configuration.
A rollout is counted as successful when all three wiping strokes are completed.

\subsection{Training Setup}
\begin{table}[t]
    \centering
    \caption{Training configuration, identical across the four configurations except for the two ablated factors.}
    \label{tab:hyperparams}
    \small
    \begin{tabular*}{\columnwidth}{@{\extracolsep{\fill}}lc}
        \toprule
        Parameter & Value \\
        \midrule
        Encoder layers & 4 \\
        Decoder layers & 4 \\
        Hidden dimension & 512 \\
        Feedforward dimension & 2048 \\
        Attention heads & 8 \\
        Latent dimension ($\bm{z}_c$ / $\bm{z}_u$) & 2 / 1 \\
        Modifier head hidden widths & [3, 3] \\
        Optimizer & AdamW~\cite{loshchilov2019decoupled} \\
        Learning rate & $1\times10^{-4}$ \\
        Loss weights ($\lambda_\mathrm{rec}$, $\lambda_\mathrm{kl}$, $\lambda_\mathrm{modi}$) & 1.0, 0.3, 2.5 \\
        \bottomrule
    \end{tabular*}
\end{table}

Joint angles, angular velocities, and torques were recorded from both robots via the four-channel bilateral control system at 1000 Hz, yielding 15-dimensional joint data each.
Following~\cite{oishi2025imitation}, the models take proprioceptive state alone as input, so the comparison isolates the effect of the conditioning mechanism from visual factors.
Table~\ref{tab:hyperparams} lists the training configuration.
The latent commands used at inference were obtained from each trained model separately, so that every configuration is evaluated with commands drawn from its own latent space.

\subsection{Ablation Design}\label{sec:ablation}

To isolate the contribution of modifier-conditioned decoding, we conducted a $2 \times 2$ ablation over two factors.

\textbf{Factor A (adaLN-Zero).} Whether the constrained latent $\bm{z}_c$ additionally modulates the LayerNorm parameters of every decoder sub-layer.

\textbf{Factor B ($\bm{z}_u$).} Whether the unconstrained latent $\bm{z}_u$ is included in the latent token supplied to the encoder.
This component originates in the conditional variational formulation of ACT~\cite{zhao2023learning}, where a style variable absorbs the variability of human demonstrations so that the decoder need not learn a deterministic map, and it is sampled from the posterior during training and fixed to the prior mean at inference.
Unlike the constrained latent, whose value is supplied explicitly at inference, this leaves a mismatch between the two phases, which reaches the decoder through the encoder output.

All four configurations condition the encoder by appending a modifier token to its input, share an identical latent structure and training objective, and differ only in whether $\bm{z}_c$ additionally reaches the decoder through adaLN-Zero and whether $\bm{z}_u$ is supplied.
The configuration with adaLN-Zero disabled and $\bm{z}_u$ enabled reproduces the action chunking baseline of~\cite{oishi2025imitation}, and Bi-MoDe enables adaLN-Zero and removes $\bm{z}_u$.
The remaining two configurations each apply one of these changes in isolation.

\subsection{Evaluation Metrics}\label{sec:metrics}

Directive-following fidelity is quantified by the Modifier Directive Error (MDE), following~\cite{oishi2025imitation} with one modification described below.
For each modifier type, a task-relevant scalar feature is extracted from every trial.
The temporal feature is the average time required to complete the three wiping cycles.
The physical feature is obtained by detecting the three wiping cycles, taking the median of the estimated torque within each cycle for Joint2 and Joint3, and averaging these medians into a single value per trial.

Trials are grouped by the commanded level $x \in \{0.0,\, 0.5,\, 1.0\}$ of the modifier under evaluation, pooling across the levels of the other modifier, so that each of the three groups contains 15 trials.
A line $y = cx + d$ is fitted by unweighted least squares to the three group means, and the same procedure applied to the demonstrations yields a reference line $y = ax + b$.
MDE is the distance between the two lines in coefficient space, with both terms normalized by a common constant $R$:
\begin{equation}
    \mathrm{MDE} = \sqrt{
        \left( \frac{a-c}{R} \right)^{2} +
        \left( \frac{b-d}{R} \right)^{2}
    }
    \label{eq:mde}
\end{equation}
where $R$ is the range of the individual reference features pooled across all three levels, computed from individual trials rather than from group means so that it reflects the full spread of the demonstrated behavior; here $R = 4.643$~s on the temporal axis and $R = 0.830$~Nm on the physical axis.
Lower MDE indicates closer agreement between the intended directive and the generated motion.

The original formulation replaces $R$ with the reference coefficients $a$ and $b$ individually, which is ill-conditioned for the physical axis of a contact task: the weakest force level is contact that is barely maintained, so the reference intercept lies close to zero by construction ($a = 0.679$, $b = 0.024$~Nm) and dominates the metric, contributing $0.875$ against $0.032$ from the slope term for Bi-MoDe.
Normalizing both terms by $R$ removes this sensitivity and expresses the error in units of the behavioral range the demonstrations span.

We additionally report the \textit{slope ratio} $c/a$, which isolates the slope component of Eq.~\eqref{eq:mde} and expresses how strongly the directive modulates the realized behavior.
A slope ratio of $1.0$ indicates that the generated motion spans the same dynamic range as the demonstrations, whereas a value near $0$ indicates that the directive is effectively ignored.
The slope ratio contains no intercept term and is therefore unaffected by the choice of normalization discussed above.

\begin{figure*}[t]
\centering
\includegraphics[keepaspectratio, width=\linewidth]{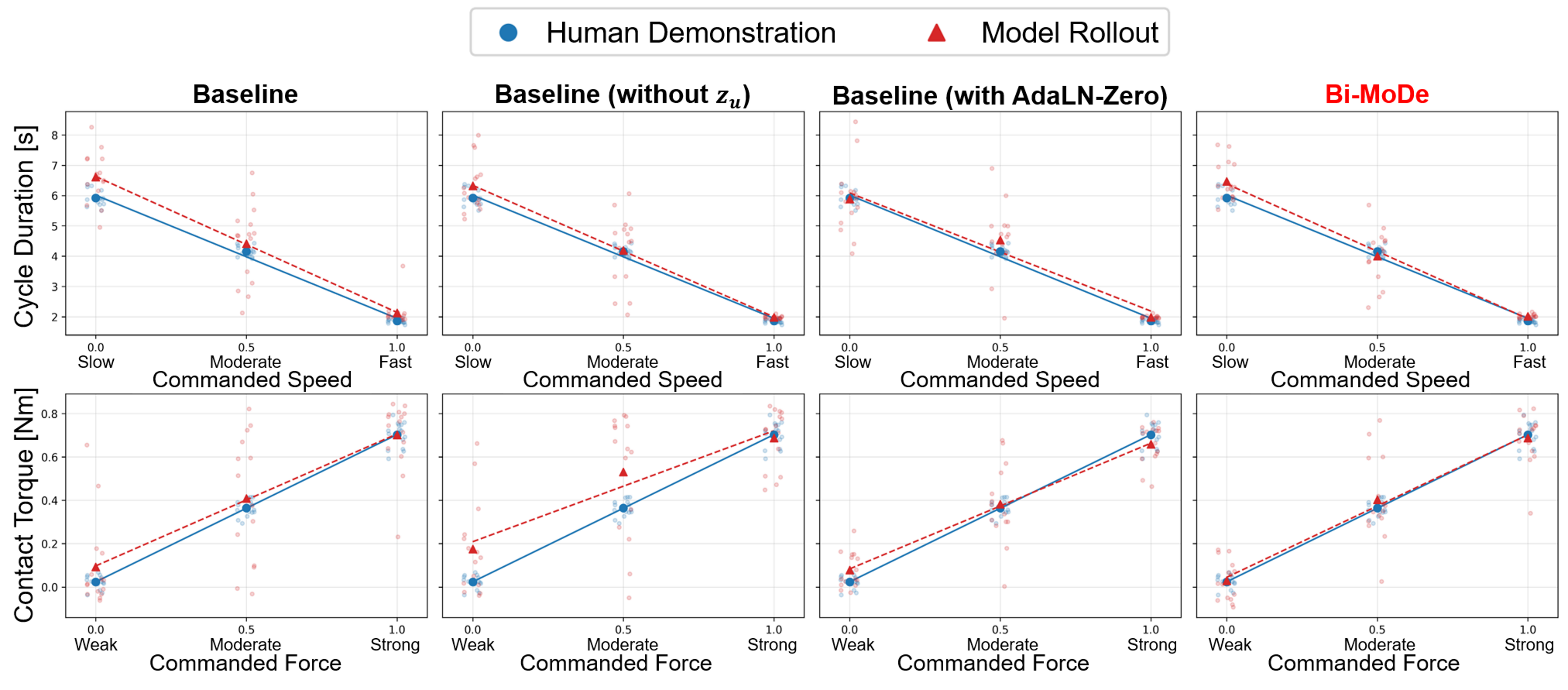}
\caption{Cycle duration versus commanded speed (top) and contact torque versus commanded force (bottom). Columns are the four configurations of Table~\ref{tab:mde} from left to right, each with the reference line fitted to the demonstrations.}
\label{fig:adherence}
\end{figure*}

\begin{figure*}[t]
\centering
\includegraphics[keepaspectratio, width=\linewidth]{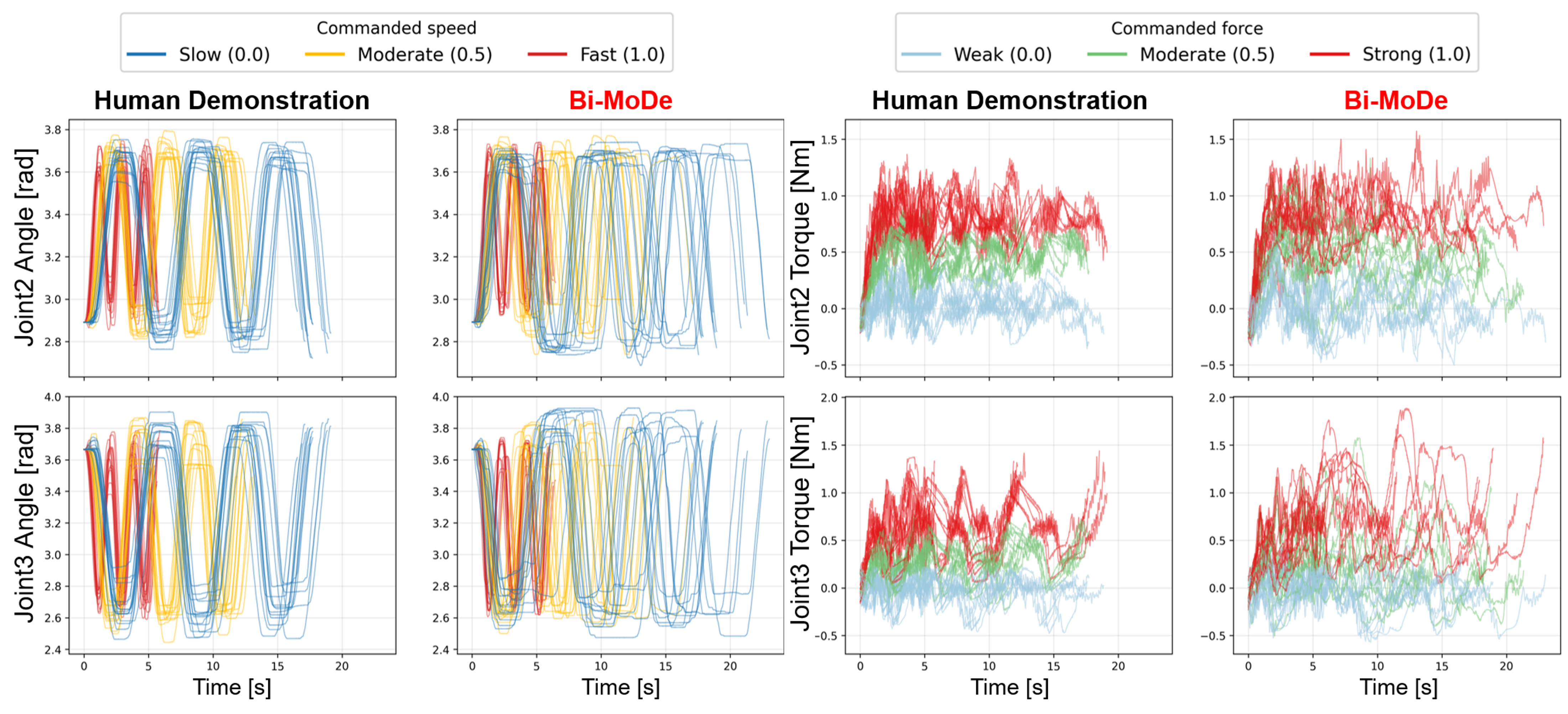}
\caption{Joint2 and Joint3 profiles, with the demonstrations on the left and Bi-MoDe on the right. Angles are shown for the three commanded speed levels and torques for the three commanded force levels.}
\label{fig:profile}
\end{figure*}

\begin{table}[t]
\centering
\caption{Task success rate and directive-following performance.
MDE $\downarrow$; slope ratio $\rightarrow 1.0$.}
\label{tab:mde}

\resizebox{\columnwidth}{!}{
\begin{tabular}{lcc c cc cc}
\toprule
\multirow{2}{*}{Configuration}
& \multicolumn{2}{c}{Design}
& \multirow{2}{*}{TSR}
& \multicolumn{2}{c}{Physical}
& \multicolumn{2}{c}{Temporal} \\
\cmidrule(lr){2-3}
\cmidrule(lr){5-6}
\cmidrule(lr){7-8}
& Decoder cond. & $\bm{z}_u$
& & MDE & slope & MDE & slope \\
\midrule

Baseline (ACT-based)~\cite{oishi2025imitation}
& -- & \checkmark
& 45/45 & 0.123 & 0.896 & 0.161 & 1.104 \\

No $\bm{z}_u$
& -- & --
& 45/45 & 0.300 & 0.754 & 0.093 & 1.069 \\

Decoder conditioned
& \checkmark & \checkmark
& 45/45 & 0.139 & 0.854 & 0.039 & 0.959 \\

\textbf{Bi-MoDe (Ours)}
& \checkmark & --
& 45/45 & \textbf{0.036} & \textbf{0.968} & 0.116 & 1.096 \\

\bottomrule
\end{tabular}
}
\end{table}

\subsection{Results}
\subsubsection{Overall}
Every rollout of every configuration completed the three wiping strokes, so the task success rate (TSR) is $45/45$ throughout and does not separate the configurations; the comparison below therefore concerns how faithfully the commanded level is realized rather than whether the task is finished.
Fig.~\ref{fig:adherence} plots the realized feature against the commanded level for both axes; points are group means over 15 trials and lines are least-squares fits, so that MDE is the distance between each generated line and the reference line fitted to the demonstrations.
The temporal panel slopes downward because a higher commanded level means a shorter stroke.
Table~\ref{tab:mde} summarizes the quantitative results.

Fig.~\ref{fig:profile} shows the joint angle and torque profiles of Joint2 and Joint3, with the demonstrations on the left and Bi-MoDe on the right.
The angle profiles are drawn for the three commanded speed levels and the torque profiles for the three commanded force levels.
In both halves the angle traces separate in duration while following the same path, and the torque traces separate in magnitude during the contact phase, so each directive acts on the quantity it is meant to control.
The separation Bi-MoDe produces on the force axis follows that present in the demonstrations, which is what the physical slope ratio of $0.968$ measures.

\subsubsection{Physical directive following}
The two design factors interact.
Reading the slope ratio, whose distance from unity measures how much of the demonstrated force range the policy reproduces, Baseline sits at $0.896$; removing $\bm{z}_u$ alone moves it to $0.754$ and introducing adaLN-Zero while retaining $\bm{z}_u$ moves it to $0.854$, both further from unity, whereas Bi-MoDe, which combines the two, reaches $0.968$.
MDE follows the same ordering, rising from $0.123$ to $0.300$ and to $0.139$ under either single change and falling to $0.036$ under both.
The agreement between the two metrics is informative because the slope ratio is computed without reference to the intercept, so the pattern is a property of the configurations rather than of how the error is normalized.

We interpret the pattern as follows.
The two components of the latent reach the decoder by different routes.
The constrained latent is supplied to the encoder and, in the proposed configuration, also modulates every decoder layer; the unconstrained latent enters only through the encoder, so whatever it carries arrives mixed into the encoder output.
Because $\bm{z}_u$ is sampled from the posterior during training and fixed to the prior mean at inference, that mixture differs between the two phases, and the directive must be recovered from it.
When the decoder has no direct access to the directive, this is tolerable, since the residual capacity that $\bm{z}_u$ provides also absorbs demonstration-level variability that the reconstruction objective would otherwise have to explain.
Once adaLN-Zero supplies the directive to the decoder directly, that capacity is no longer needed for the directive, while the mismatch it introduces into the encoder output remains, and removing it improves directive following.
Neither change helps alone because each removes one of the two effects without the other.

\subsubsection{Temporal directive following}
All four configurations follow the temporal directive closely.
Every slope ratio lies roughly within ten percent of unity, between $0.959$ and $1.104$, and every MDE lies between $0.039$ and $0.161$, so the generated motion spans essentially the full range of stroke durations present in the demonstrations regardless of how the modifier reaches the decoder.
The differences among the four are small and do not follow the pattern observed on the physical axis, and we therefore do not order the configurations here.
The temporal modifier acts on the duration of the wiping stroke, a quantity the policy can regulate through the commanded joint trajectory alone, whereas the physical modifier must be realized through contact with the board and is therefore mediated by the interaction with the surface.
That the temporal axis is followed accurately by every configuration while the physical axis separates them indicates that the conditioning design matters where the directive must survive contact.

\section{CONCLUSIONS}

This paper addressed the problem of specifying how a manipulation task should be executed, rather than which task to execute, within a bilateral control-based imitation learning framework.
We introduced Bi-MoDe, which propagates a constrained latent aligned with scalar modifier directives through every layer of the Transformer action decoder via adaLN-Zero.
In the real-world contact-rich wiping task, Bi-MoDe improved physical directive following while maintaining comparable temporal modulation.
A $2 \times 2$ ablation further showed that decoder conditioning and latent-space composition jointly contribute to physical directive following.

The present evaluation is limited to a single contact-rich task and to the directive levels represented in the training data.
Future work will evaluate Bi-MoDe across diverse contact-rich manipulation tasks involving different types of physical interaction and robotic platforms, and investigate its generalization to unseen directive levels and task conditions.

\end{document}